\documentclass[10pt,twocolumn,letterpaper]{article}

\usepackage{iccv}              

\definecolor{iccvblue}{rgb}{0.21,0.49,0.74}
\usepackage[pagebackref,breaklinks,colorlinks,allcolors=iccvblue]{hyperref}

\def\paperID{5256} 
\def\confName{ICCV}
\def\confYear{2025}

\title{What You Have is What You Track: Adaptive and Robust Multimodal Tracking

}

\usepackage{multirow}
\usepackage{makecell}
\usepackage{supertabular}
\usepackage{colortbl}
\usepackage{svg}
\usepackage[ruled,linesnumbered]{algorithm2e}
\usepackage{booktabs}  
\usepackage{siunitx}     
\usepackage{multirow}    
\usepackage{array}       
\usepackage{booktabs}
\usepackage{pifont}
\usepackage{xcolor}
\usepackage{hyperref}

\author{Yuedong Tan$^{1,2,3}$\quad Jiawei Shao$^{1}\thanks{Corresponding}$  \quad Eduard Zamfir$^{2}$  \quad  Ruanjun Li$^{4}$   \quad Zhaochong An$^{5}$
      \\  Chao Ma$^{6}$  \quad Danda Paudel$^{3}$ \quad Luc Van Gool$^{3}$ \quad Radu Timofte$^{2}$  \quad Zongwei Wu$^{2} \footnotemark[1]$
    \\ \small
$^{1}$ TeleAI, China Telecom \quad  $^{2}$ Computer Vision Lab, CAIDAS \& IFI, University of Wurzburg  \quad 
        $^3$ INSAIT, Sofia University  \\ \small
        $^4$ ShanghaiTech University \quad  $^5$ University of Copenhagen  \quad  $^6$ AI Institute, Shanghai Jiao Tong University     }

\begin{document}
\maketitle
\begin{abstract}
Multimodal data is known to be helpful for visual tracking by improving robustness to appearance variations. However, sensor synchronization challenges often compromise data availability, particularly in video settings where shortages can be temporal. Despite its importance, this area remains underexplored. In this paper, we present the first comprehensive study on tracker performance with temporally incomplete multimodal data. Unsurprisingly, under such a circumstance, existing trackers exhibit significant performance degradation, as their rigid architectures lack the adaptability needed to effectively handle missing modalities.
To address these limitations, we propose a flexible framework for robust multimodal tracking. We venture that a tracker should dynamically activate computational units based on missing data rates. This is achieved through a novel Heterogeneous Mixture-of-Experts fusion mechanism with adaptive complexity, coupled with a video-level masking strategy that ensures both temporal consistency and spatial completeness — critical for effective video tracking. Surprisingly, our model not only adapts to varying missing rates but also adjusts to scene complexity. Extensive experiments show that our model achieves SOTA performance across 9 benchmarks, excelling in both conventional complete and missing modality settings. The code and benchmark will be publicly available at \href{https://github.com/supertyd/FlexTrack/tree/main}{here}.
\end{abstract}

\section{Introduction}
\label{sec:intro}

\begin{figure}[ht]
    \centering
    
    \begin{subfigure}[b]{\columnwidth}
        \includegraphics[width=\linewidth]{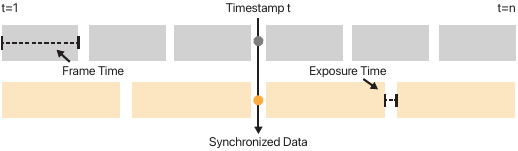}
        \caption{\textit{Sensor Synchronization.}}
        \label{fig:a}
    \end{subfigure}%
    
    \begin{subfigure}[b]{0.49\columnwidth}
        \includegraphics[width=\linewidth]{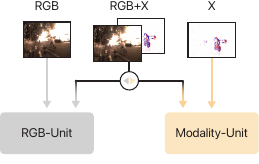}
        \caption{\textit{Prior Work.}}
        \label{fig:b}
    \end{subfigure}
    \hfill
    \begin{subfigure}[b]{0.49\columnwidth}
        \includegraphics[width=\linewidth]{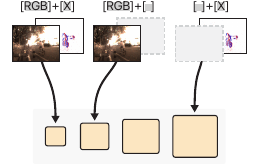}
        \caption{\textit{Heterogenous Experts.}}
        \label{fig:c}
    \end{subfigure}
    \vspace{-3mm}
\caption{Synchronization has long been a challenge in multisensor fusion. As illustrated in (a), differences in frame timing and exposure duration make it difficult for a multisensor system to obtain perfectly synchronized pairs, often resulting in missing data. To address this issue, existing trackers  \cite{ipt}, as shown in (b), handle each missing-data scenario individually, employing tailored and individual strategies for different cases. However, this approach does not account for the complexity variations introduced by different types of missing data. In contrast, in (c), we take a more holistic approach, unifying both missing and full-modality scenarios within a single framework. Our method leverages Heterogeneous Experts with adaptive complexity, providing a personalized yet structurally unified solution for robust tracking.}
\vspace{-6mm}
\label{fig:teaser}
\end{figure}

Visual object tracking aims to localize a target object in a video while handling dynamic appearance changes \cite{sparsett,artrack,romtrack,seqtrack,swintrack,keeptrack}. To improve robustness against occlusions, illumination shifts, and environmental variations, multimodal tracking has emerged as a powerful solution \cite{rgbe1,rgbt234,wang2024event}. Complementary information from multiple sensors (e.g., depth, thermal, event) has shown the effectiveness of improving unimodal approaches under ideal conditions.

However, real-world multimodal data is rarely perfect \cite{missing_1,missing_2,missing_3}. Sensors often suffer from synchronization failures and temporary dropouts. As illustrated in Fig.~\ref{fig:teaser}, the inherent differences in exposure time and frame rates make it extremely challenging to achieve perfect synchronization across multiple sensors. Existing tracking datasets \cite{rgbt234,li2021lasher,depthtrack} simplify this issue by considering only paired modalities, such as RGB and depth. However, this setting remains limited, as it does not generalize to real-world scenarios where multiple modalities may be intermittently unavailable within a given temporal window during video tracking. Despite its practical importance, such a missing scenario has not been well-adjusted.

In this paper, we conduct a comprehensive study on temporally missing modalities in multimodal tracking \cite{hou2024sdstrack,vipt,chen2024sutrack}. For each benchmark dataset, we introduce a missing-data variation that simulates real-world synchronization challenges by incorporating diverse missing patterns, excluding cases where all modalities are absent simultaneously. Benchmarking existing methods under this setting reveals that they struggle to maintain reliable performance, primarily due to their reliance on template-matching frameworks between a reference template and a search region~\cite{ostrack,vipt,swintrack,seqtrack,siambag,siamcar}, which are not designed to handle missing modalities. A recent attempt \cite{ipt} to address this issue introduces placeholder and tailored prompts for both complete and missing cases. However, its rigid design raises a fundamental question: \textit{Should a tracker maintain the same computational complexity regardless of whether modalities are missing or available?}

Intuitively, we argue that an effective tracker should dynamically adapt, adjusting the computational resources it uses based on the available information and the severity of missing data. To achieve this, we propose \textbf{FlexTrack}, a unified framework designed to robustly handle multimodal tracking in both full and missing modality scenarios. Technically, we introduce a novel heterogeneous Mixture-of-Experts(HMoE) mechanism. This approach leverages varying levels of complexity and offers a unique advantage by routing and activating the most relevant experts based on the current data. Additionally, we propose a novel video-level masking strategy that enforces temporal consistency at the clip level while preserving contextual completeness, thus facilitating more robust object tracking.

Our tracker stands out from SOTA counterparts, as demonstrated by the holistic comparison in Tab.~\ref{tab:comparison}, making it more adaptable to tracking with available data. Extensive evaluations on both conventional benchmarks with full modalities and our newly created missing-data variations validate its effectiveness. Notebly, on full-modality benchmarks, we outperform the current SOTA by 2.6\%, with the performance gap widening significantly under the missing-data setting, where we achieve a margin of 10.2\%.

\newcommand{\greencheck}{\textcolor{black}{\ding{51}}}
\newcommand{\redcross}{\textcolor{red}{\ding{55}}}
\begin{table}[t]
    \centering
    \caption{Comparison of SOTA multimodal tracking models.}
        \vspace{-2mm}
    \resizebox{\columnwidth}{!}{
    \begin{tabular}{c|cccc}
        \toprule
        \textbf{Method} & \textbf{Unified Arch.} & \textbf{Unified Param.} & \textbf{Missing Modal} & \textbf{Flexi-Modal Adaptive} \\
        \midrule
        IPT \cite{ipt} & \redcross & \redcross & \greencheck  & \redcross\\
        ViPT \cite{vipt} & \greencheck & \redcross & \redcross & \redcross \\
        UnTrack \cite{untrack} & \greencheck & \greencheck & \redcross & \redcross \\
        \rowcolor{gray!20}
        Ours & \greencheck & \greencheck & \greencheck & \greencheck \\
        \bottomrule 
    \end{tabular}
    }
    \label{tab:comparison}
    \vspace{-4mm}
\end{table}

\section{Related Work}
\label{sec:related_work}

\begin{figure*}[t]
\centering
\includegraphics[width=\linewidth,keepaspectratio]{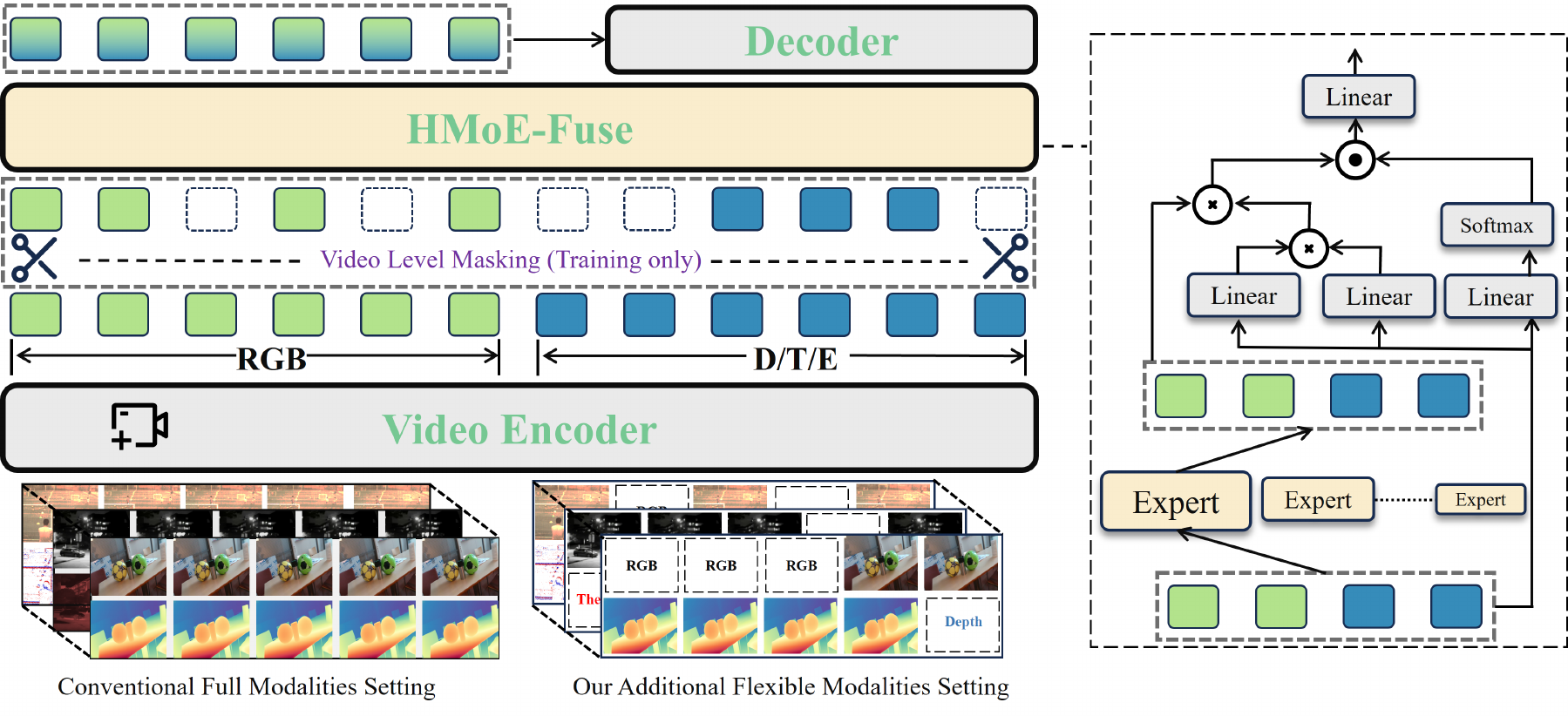}
\vspace{-7mm}
\caption{\textbf{Our framework:} Conventional trackers operate under the strong assumption that all frames are always available, disregarding practical scenarios with missing modalities. To address this limitation, we introduce heterogeneous experts, which dynamically switch between different experts to adapt \textit{test-time complexity}. To further strengthen the robust learning, we employ a video-level masking strategy during training, encouraging the model to capture temporal continuity despite missing data.}
\vspace{-3mm}
\label{fig:framework}
\end{figure*}

\subsection{Multimodal Tracking}
Modern object tracking has advanced significantly with the adoption of Siamese networks and transformer architectures\cite{gao2022aiatrack,ostrack,seqtrack,artrack}. 
Despite these advancements, tracking remains constrained by the limitations of single-modal (e.g., RGB) inputs, which struggle with challenges like occlusion or low-light conditions \cite{wang2023visevent}.
Multimodal tracking \cite{rgbt234,rgbd1k,sttrack}, which integrates RGB data with additional modalities (e.g., depth, thermal, event), has proven effective, especially when temporal cues are exploited \cite{hou2024sdstrack,hong2024onetracker,untrack,vipt}.
 However, these methods assume consistent availability of modality, which is unrealistic in real world scenarios \cite{ipt}. Our work addresses this gap by introducing a video-level masking strategy that explicitly handles missing modalities while retaining video-level temporal coherence, 
 and a HMoE mechanism to switch different sizes of experts to handle different levels of modality missing.

\subsection{Missing Modalities}
Missing modalities \cite{missing_1,missing_2,missing_3} is a wildly 
researched problem. Real-world multimodal systems often face the challenge of having to handle cases where certain data modalities may be missing or incomplete. Existing methods for addressing missing modalities in object tracking include prompt-based approaches and generative models such as GANs \cite{gans} or diffusion models \cite{croitoru2023diffusion,missing_diff}. While effective, these methods often introduce additional computational overhead. Alternatively, some approaches employ distillation techniques\cite{missing_survey,missing_disstilation}, utilizing full-modal models to distill models with missing modalities. However, this can result in reduced model flexibility. Another common strategy involves actively masking modalities during training to enhance the model's robustness to each modality \cite{missing_reconstraction}. This technique typically involves randomly masking different modalities to improve the model's resilience. However, existing models primarily focus on single-instance prediction tasks such as classification and detection. In contrast, our model explores a novel video-level masking strategy for object tracking, effectively leveraging temporal information to mitigate the impact of missing modalities.

\subsection{Mixture-of-Experts}
Mixture-of-Experts(MoE) models utilize multiple specialized expert networks to adaptively select and process inputs, thereby enhancing performance across various tasks. Notably, large-scale MoE implementations such as DeepSeekMoE \cite{dai2024deepseekmoe}, and Llama-MoE \cite{llama-moe} have demonstrated exceptional efficiency in performance. Beyond their prominence in large language models, MoE architectures have also been effectively applied in multimodal fusion, leveraging flexible expert switching to accommodate diverse scenarios. For instance, FuseMoE \cite{han2025fusemoe} and Flex-MoE \cite{yunflex-moe} employ MoE to integrate different modalities and address issues arising from missing modalities. However, they all have the uniform design for the expert size.

In contrast, our approach introduces a novel module by combining temporal information with experts of varying sizes to address the challenges of adaptive fusion. This methodology enables more precise and context-aware processing of multimodal data, thereby enhancing the model's robustness and flexibility in handling diverse scenarios.

\section{Method}

\subsection{Overview}

\begin{figure*}[t]
\centering
\includegraphics[width=\linewidth,keepaspectratio]{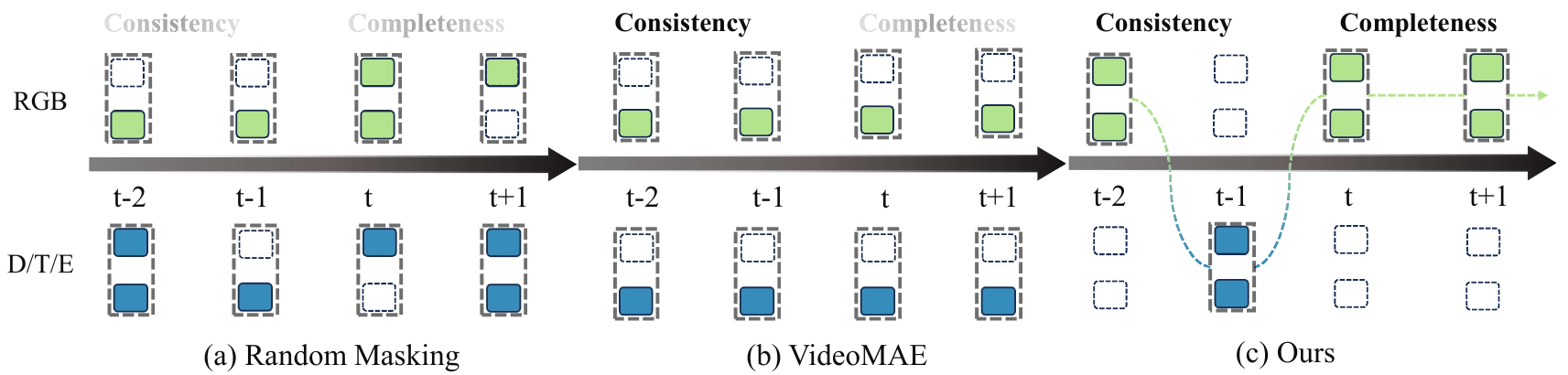}
\vspace{-7mm}
\caption{Comparison of different masking strategies for multimodal settings. (a) Conventional random or MAE-like masking \cite{wu2023dropmae} can be easily extended to multimodal settings but disrupts both spatial completeness and temporal continuity in video frames. (b) Extending VideoMAE \cite{tong2022videomae} to multimodal settings enhances temporal consistency by preserving token continuity within each modality. However, it applies the same spatial randomness across all frames, resulting in persistent spatial information loss. (c) We argue that both spatial completeness and temporal consistency are crucial for object tracking: spatial completeness provides a detailed understanding of appearance, while temporal consistency captures dynamic changes. To achieve this balance, we propose a structured, video-level masking strategy that ensures the full spatial representation remains consistently retrievable in available data across different timestamps.}
\vspace{-4mm}
\label{fig:mask-strategy}
\end{figure*}

In this paper, we propose a unified framework for object tracking that can handle flexible combinations of modalities. As illustrated in Fig.~\ref{fig:framework}, our framework comprises two primary components: a multimodal fusion module and a video-level masking strategy which is only used in the training stage. Our tracker accepts video clips and search regions from RGB and auxiliary modalities (depth, thermal, and event modalities, collectively referred to as \textbf{X}) as input.

To address the challenge of varying modality completeness in tracking, we propose the Heterogeneous Mixture-of-Experts Fusion (HMoE-Fuse) module. Unlike conventional approaches that statically allocate computational resources, we dynamically adjust the model's test-time complexity based on the availability of multimodal inputs.

Additionally, to enhance temporal continuity learning under partial modality-missing conditions, we propose a multimodal video-level masking mechanism during training. In each time-step, we keep at least one frame available to focus on learning spatio-temporal consistency.

\subsection{Heterogeneous Mixture-of-Experts Fusion}
As illustrated in Fig.~\ref{fig:framework}, we first feed both the RGB and X frames into the video encoder, which produces several token representations: RGB search region tokens, denoted as $T_{\text{rgb}}^{s}$; video clip tokens ranging from $T_{\text{rgb}}^{c1}$ to $T_{\text{rgb}}^{ci}$; and, for additional X, we obtain  $T_{x}^{s}$, $T_{x}^{c1}$ to $T_{x}^{ci}$, where $i \in {1, \dots, N}$.

First, we concatenate all tokens to form the sequence token $T_{v}$. The core component of the HMoE-Fuse is a heterogeneous MoE, which differs from typical MoEs where each expert is constructed with a distinct architecture. Specifically, each expert in our model consists of a linear layer, and its hidden state dimension is set as $2^d$, where $d\in \{2, \dots,D-1\}$.

Contrary to prior MoEs \cite{dai2024deepseekmoe,han2025fusemoe}, we route the entire video clips to experts instead of individual tokens or images.
First, a gating function $G$ produces a weight distribution, where higher values indicate better input-expert alignment.  The output tokens $T_{y}^1$ of the HMoE layer is the weighted sum of outputs from the top-$K$ activated experts:
\begin{equation}
\begin{gathered}
    g_n =
    \begin{cases}
        \text{Softmax}(G(T_{v}))_n, \quad \text{if } G(T_{v})_n \in \text{Top-K}(G(T_{v}) ) \\
        0, \quad \text{otherwise}
    \end{cases} \\
    T_{y}^1 = \sum_{n=1}^{M} g_n E_n(T_{v}).
\end{gathered}
\end{equation}
where $g_n$ represents the gating values derived from the function $G(T_v)$, and $E_n(T_{v})$ is the output of the $n$-th expert. Besides, the gating function ensures that only the top $K$ experts are activated, here we set $K$=2 in our method. With consistent video frame input, our HMoE selects the most appropriate experts to deal with missing modality cases.
Simultaneously, we process the tokens $T_{v}$ through two linear layers, yielding an intermediate transformation:
\begin{equation}
T_{y}^2 = T_{v}W_1(T_{v}W_2)^T,
\end{equation}
where $W_1$ and $W_2$ are learned weight matrices that perform linear attention on the tokens.

Next, we perform a further transformation on the tokens by combining the expert outputs and the intermediate result:
\begin{equation}
T_{y}^3 = T_{y}^1(T_{y}^2)^T.
\end{equation}
This operation integrates the expert outputs with the transformed tokens $T_y^2$.
Finally, we apply a softmax operation and a linear transformation to further refine the output:
\begin{equation}
T_{y}^4 = \text{Softmax}(T_vW_3)T_y^3,
\end{equation}
and
\begin{equation}
T_{y}^5 = T_{y}^4W_4,
\end{equation}
where $W_3$ and $W_4$ are additional learnable weight matrices. The softmax operation is applied to $T_vW_3$ to scale the transformed tokens, and $W_4$ further adjusts the resulting values.

\subsection{Multimodal Video-Level Masking Strategy}
During inference, we find that our multimodal tracker fail to handle the missing modality for its substantial dependence on each frame in the video clip. 
 (For a correspondence experiment, please refer to Tab.~\ref{tab:masking_variation})

While VideoMAE~\cite{tong2022videomae}'s tube masking strategy demonstrates effectiveness in multiple video tasks through video cube masking, its design fundamentally assumes access to subsequent frames for masked content inference. However, the assumption is invalid for real-time tracking as future frames are inaccessible. Specifically, models cannot utilize future temporal information to compensate for current modality absence, which poses a critical challenge in streaming tracking scenarios. 
We identify a unique advantage in multimodal tracking: temporal continuity persists even under spatial dropout in a modality. While individual frames may suffer from partial modality absence (e.g., missing depth data in frame $t$), adjacent frames ($t-\delta$ to $t+\delta$) inherently preserve temporal coherence through at least one persistent modality information. As shown in Fig.~\ref{fig:mask-strategy}, we exploit this through a multimodal video-level masking strategy to enforce temporal modeling.

\newlength{\textfloatsepsave} \setlength{\textfloatsepsave}{\textfloatsep} \setlength{\textfloatsep}{0pt}
\begin{algorithm}[t]
\SetNoFillComment
\SetArgSty{textnormal}
\KwIn{RGB and X search region tokens $T_{rgb}^s, T_{x}^s$, RGB and X video clip tokens $T_{rgb}^{ci}, T_{x}^{ci}$, $i \in \{1, \dots, N\}$, batch size $B$, video clip mask ratio $\alpha$.}
\KwOut{Updated tokens with applied video-level masking.}

\For{$b = 1$ \KwTo $B$}{
    $M_s \gets \text{random}([[1,1], [1,1], [1,1], [1,0], [0,1]])$\;
    $T_x^s \gets M_s[0] \times T_x^s$\;
    $T_{rgb}^s \gets M_s[1] \times T_{rgb}^s$\;
    
$p \sim \mathcal{U}(0, 1)$ \tcp*{Randomly select $X$ from the interval [0, 1];} 
    \If{$p < \alpha$}{
        \For{$i = 1$ \KwTo $N$}{
            $M_{vi} \gets \text{random}([[1,1], [1,0], [0,1]])$\;
            $T_x^{ci} \gets M_{vi}[0] \times T_x^{ci}$\;
            $T_{rgb}^{ci} \gets M_{vi}[1] \times T_{rgb}^{ci}$\;
        }
    }
}
\caption{Video-level masking}
\label{alg:mask}
\end{algorithm}

\vspace{-0.6mm}

We illustrate the proposed video-level masking strategy in Algorithm~\ref{alg:mask}. During training, we first apply search region masking by randomly selecting one of five predefined modality patterns to partially mask the RGB/X tokens in the search region. After applying the search region mask, a probability $p$ is sampled from a uniform distribution over the interval [0, 1].
If the value of $p$ falls below the predetermined threshold of $\alpha$, this signifies the conditions necessary for implementing video-level masking.
Thus, the algorithm proceeds to generate individual clip-level masks $M_{vi},i \in{1,\dots,N}$ for each of the video clips.

These masks are applied to the corresponding tokens of both modalities, further refining the focus of the model on the most pertinent information.

\subsection{Head and Loss Function}
 To align with current multimodal trackers~\cite{vipt,hou2024sdstrack,hong2024onetracker,untrack}, we utilize both a classification head and a regression head to localize the object center and approximate its height and width. Consequently, we adopt the same loss functions as these trackers, specifically the classification loss $\mathcal{L}_{cls}$, the L1 loss $\mathcal{L}_{l1}$, and the GIoU loss $\mathcal{L}_{GIoU}$.
Beyond the standard tracking task losses, we integrate the importance loss $\mathcal{L}_{important}$ and balance loss $\mathcal{L}_{balance}$ from the Switch Transformer~\cite{fedus2022switch} to ensure load balancing among experts and prevent imbalance within a single expert. The auxiliary loss $\mathcal{L}_{aux}$ is defined as follows:
\begin{equation}
\mathcal{L}_{aux} = \mathcal{L}_{balance} + \mathcal{L}_{important}.
\end{equation}

The overall loss function is formulated as:
\begin{equation}
\mathcal{L}_{all} = \lambda_1 \mathcal{L}_{aux} + \lambda_2 \mathcal{L}_{cls} + \lambda_3 \mathcal{L}_{l1} + \lambda_4 \mathcal{L}_{GIoU},
\end{equation}
where $\mathcal{L}_{all}$ represents the total loss used in the training process of our method. $\lambda_1$, $\lambda_2$, $\lambda_3$, $\lambda_4$ is the hyperparameter with default values $\lambda_1$ = 1, $\lambda_2$ = 5, $\lambda_3$ = 2, and $\lambda_4$ = 1.

\section{Experiment}

\begin{table}[t]\normalsize
    \caption{SOTA comparisons on RGB-Event tracking dataset.
    }
\label{tab-sota-rgbe}
\vspace{-2mm}
  \centering
\resizebox{1\linewidth}{!}{
  \setlength{\tabcolsep}{6mm}{
    \small
    \begin{tabular}{l|cc}
    \toprule
    \multirow{2}*{Method} & \multicolumn{2}{c}{VisEvent~\cite{wang2023visevent}}\\
        \cline{2-3}
 & P& AUC\\
    \midrule[0.5pt]

        FlexTrack &\textbf{\textcolor{red}{81.4}} &\textbf{\textcolor{red}{64.1}}\\

        \midrule[0.1pt]

STTrack~\cite{sttrack}&78.6 &61.9    \\
SUTrack~\cite{chen2024sutrack} &79.9 &62.7 \\

OneTracker~\cite{hong2024onetracker}&76.7 &60.8\\

SDSTrack~\cite{hou2024sdstrack} &76.7 &59.7\\

UnTrack~\cite{untrack}&75.5 &58.9 \\
ViPT~\cite{vipt}&75.8 &59.2\\
 
ProTrack~\cite{protrack}&63.2 &47.1\\

MCITrack~\cite{mcitrack} &69.7  &53.3 \\
OSTrack~\cite{ostrack} &69.5&53.4 \\
SiamRCNN\_E~\cite{SiamRCNN} &65.9 &49.9 \\
TransT\_E~\cite{transt} &65.0 &47.4 \\
LTMU\_E~\cite{LTMU} &65.5&45.9 \\
PrDiMP\_E~\cite{PrDiMP} &64.4 &45.3 \\
VITAL\_E~\cite{VITAL} &64.9&41.5 \\
MDNet\_E~\cite{MDNet} &66.1&42.6 \\
ATOM\_E~\cite{atom} &60.8 &41.2 \\
SiamCar\_E~\cite{Stark} &59.9 &42.0 \\
SiamBAN\_E~\cite{siamban} &59.1 &40.5 \\
SiamMask\_E~\cite{SiamMask} &56.2 &36.9 \\

    \bottomrule
    \end{tabular}
    }
  }
  \label{rgbe}
  \vspace{2mm}
\end{table}

\subsection{Implementation Detail}

Consistent with recent trackers \cite{sttrack, hong2024onetracker, hou2024sdstrack}, our tracker is initialized with a pretrained encoder \cite{mcitrack,fastitpn} trained on large-scale datasets \cite{got10k, trackingnet, coco, lasot}. The model is trained on 8 4090 GPUs with a batch size of 32 using the AdamW optimizer, with a base learning rate of 3e-4. Training consists of 30 epochs, each containing 600,000 image pairs. The learning rate is reduced by a factor of 10 after 22 epochs.
For RGB-Thermal, we use LasHeR \cite{li2021lasher} for training, VisEvent \cite{wang2023visevent} for RGB-Event, and DepthTrack \cite{depthtrack} for RGB-Depth. To create a unified framework, we jointly train these three datasets in a single training process.
During inference, we disable masking operations to fully leverage all available modalities while retaining the robustness learned through training phase masking simulations.

\subsection{Evaluations on complete dataset:}
To demonstrate the performance of our tracker, we evaluated our tracker in 5 complete datasets. Under a strong assumption that all the frames are aligned well in temporal and there is not any missing modality caused by communication disruption and so on. 

\vspace{1mm}
\noindent \textbf{Evaluations on complete RGB-Event dataset:}
The VisEvent\cite{wang2023visevent} test set contains 320 videos covering a variety of challenging attributes. Our trackers achieved new SOTA results, exceeding current SOTA trackers by 1.5\% and 1.3\% in P and area-under-the-curve (AUC), respectively.

\vspace{1mm}
\noindent \textbf{Evaluations on complete RGB-Thermal dataset:}
For the RGB-Thermal tracking task, in Tab.~\ref{tab-sota-rgbe}, our FlexTrack models achieve new SOTA performance on both the LasHeR~\cite{li2021lasher} and RGBT234~\cite{rgbt234} datasets. Specifically, on the LasHeR benchmark, our FlexTrack achieves an AUC score of 62.0\%, surpassing the previous best-performing model, STTrack~\cite{sttrack}, by 1.7\%. On the RGBT234 benchmark, FlexTrack sets new SOTA score for MSR at 69.9\%.

\begin{table}[t]
    \caption{SOTA comparisons on RGB-Thermal tracking datasets.
    }
        \vspace{-5mm}

\label{tab-sota-rgbt}
\vspace{2mm}
  \centering
\resizebox{1\linewidth}{!}{
  \setlength{\tabcolsep}{2mm}{
    \small
    \begin{tabular}{l|ccccc}
    \toprule
    \multirow{2}*{Method} & \multicolumn{2}{c}{LasHeR~\cite{li2021lasher}} & & \multicolumn{2}{c}{RGBT234~\cite{rgbt234}} \\
        \cline{2-3} \cline{5-6}
 & P & AUC & &MPR &MSR \\
    \midrule[0.5pt]

        FlexTrack &\textbf{\textcolor{red}{77.3}}&\textbf{\textcolor{red}{62.0}} & &\textcolor{red}{\textbf{92.7}} &\textcolor{red}{\textbf{69.9}}\\
        \midrule[0.1pt]

STTrack~\cite{sttrack} &76.0 &60.3 & & 89.8 &66.7\\
SUTrack~\cite{chen2024sutrack} &74.5 &59.5 & &92.2 &69.5\\

OneTracker~\cite{hong2024onetracker} &67.2 &53.8 & & 85.7 &64.2\\
SDSTrack~\cite{hou2024sdstrack} &66.5&53.1 & &84.8 &62.5\\

UnTrack~\cite{untrack} &64.6&51.3 & &84.2& 62.5\\
    ViPT~\cite{vipt} &65.1 &52.5 & &83.5&61.7\\
    ProTrack~\cite{protrack} &53.8&42.0 & & 79.5& 59.9\\

MCITrack~\cite{mcitrack} &64.2 &50.2 & &79.1 &60.4 \\
OSTrack~\cite{ostrack} &51.5 &41.2 & &72.9&54.9 \\
APFNet~\cite{apfnet} &50.0&36.2 &&82.7 &57.9 \\
CMPP~\cite{cmpp} &-&- &&82.3 &57.5 \\
JMMAC~\cite{jmmac} &-&- &&79.0 &57.3 \\
CAT~\cite{cat} &45.0&31.4 &&80.4 &56.1 \\
FANet~\cite{fanet} &44.1&30.9 & &78.7&55.3 \\
mfDiMP~\cite{mfdimp} &44.7&34.3 & &64.6&42.8 \\
SGT~\cite{sgt} &36.5&25.1 & &72.0 &47.2 \\
HMFT~\cite{vtuav} &43.6&31.3 & &-&- \\
DAPNet~\cite{dapnet} &43.1&31.4 & &-&- \\
DAFNet~\cite{dafnet} &-&- &&79.6 &54.4 \\
MaCNet~\cite{macnet} &-&- &&79.0 &55.4 \\

    \bottomrule
    \end{tabular}
    }
  }
    \vspace{4mm}
  \label{tab:rgbt}
\end{table}

\vspace{1mm}
\noindent \textbf{Evaluations on complete RGB-Depth dataset:}
In comprehensive RGB-Depth datasets, our model has demonstrated exceptional performance in DepthTrack \cite{depthtrack}. Specifically, we have achieved an F-score that surpasses the previous best tracker, SUTrack \cite{chen2024sutrack}, by 1.9\%. Furthermore, despite training exclusively on the VisEvent \cite{wang2023visevent}, LasHeR \cite{li2021lasher}, and VisEvent \cite{wang2023visevent} datasets, our model exhibits strong out-of-distribution performance. In Tab.~\ref{tab-sota-rgbd}, experimental results indicate that our model also achieves impressive results on the VOT-RGBD22 benchmark.

\subsection{Evaluations on missing modality dataset:}
IPT~\cite{ipt} is the first method to address scenarios involving missing modalities. Initially, IPT was trained and tested exclusively on RGB-Thermal datasets. Subsequently, to better align with real-world applications, IPT developed a new dataset derived from the original one. This dataset includes instances of random missing, switched missing, and prolonged missing modalities. Building upon this dataset design, we expanded our evaluation to encompass multiple modalities including DepthTrack\cite{depthtrack}, VisEvent\cite{wang2023visevent}, LasHeR\cite{li2021lasher}, and RGBT234\cite{rgbt234}.

\begin{table}[t]
    \caption{SOTA comparisons on RGB-Depth tracking datasets.
    }
    \vspace{-5mm}
\label{tab-sota-rgbd}
\vspace{2mm}
  \centering
\resizebox{1\linewidth}{!}{
  \setlength{\tabcolsep}{1.5mm}{
    \small
    \begin{tabular}{l|ccc c ccc}
    \toprule
    \multirow{2}*{Method} & \multicolumn{3}{c}{DepthTrack~\cite{depthtrack}} & & \multicolumn{3}{c}{VOT-RGBD22~\cite{vot2022}} \\
        \cline{2-4} \cline{6-8}
 & F-score & Re & Pr& & EAO & Acc.& Rob. \\
    \midrule[0.5pt]
FlexTrack &\textcolor{red}{\textbf{67.0}}&\textcolor{red}{\textbf{66.9}}&\textcolor{red}{\textbf{67.1}} & &\textcolor{red}{\textbf{78.0}}&\textbf{\textcolor{red}{83.8}} &\textcolor{red}{\textbf{93.1}}\\

\midrule[0.1pt]

STTrack~\cite{sttrack} &63.3 &63.4 &63.2 & &77.6 & 82.5 & 93.7\\
SUTrack~\cite{chen2024sutrack} &65.1 &65.7 &64.5 & &76.5 & 82.8 & 91.8\\

OneTracker~\cite{hong2024onetracker} &60.9 &60.4 &60.7 & &72.7 & 81.9 & 87.2\\

SDSTrack~\cite{hou2024sdstrack} &61.4 &60.9 & 61.9 & &72.8 & 81.2 & 88.3\\
UnTrack~\cite{untrack} &61.0&60.8&61.1 & &72.1&82.0 &86.9\\
ViPT~\cite{vipt} &59.4&59.6&59.2 & &72.1&81.5 &87.1\\
ProTrack~\cite{protrack} &57.8&57.3&58.3 & &65.1&80.1&80.2\\

MCITrack~\cite{mcitrack} &53.8&54.9&52.7 & &74.3 &82.3&90.6\\
SPT~\cite{rgbd1k} &53.8&54.9&52.7 & &65.1&79.8&85.1\\
SBT-RGBD~\cite{sbt} &-&-&- & &70.8&80.9&86.4\\
OSTrack~\cite{ostrack} &52.9&52.2&53.6 & &67.6&80.3&83.3\\
DeT~\cite{depthtrack} &53.2&50.6&56.0 & &65.7&76.0&84.5\\
DMTrack~\cite{vot2022} &-&-&- & &65.8&75.8&85.1\\
DDiMP~\cite{vot2020} &48.5&56.9&50.3 & &-&-&-\\
ATCAIS~\cite{vot2020} &47.6&45.5&50.0 & &55.9&76.1&73.9\\
LTMU-B~\cite{LTMU} &46.0&41.7&51.2 & &-&-&-\\
GLGS-D~\cite{vot2020} &45.3&36.9&58.4 & &-&-&-\\
DAL~\cite{dal} &42.9&36.9&51.2 & &-&-&-\\
LTDSEd~\cite{VOT2019} &40.5&38.2&43.0 & &-&-&-\\
Siam-LTD~\cite{vot2020} &37.6&34.2&41.8 & &-&-&-\\
SiamM-Ds~\cite{VOT2019} &33.6&26.4&46.3 & &-&-&-\\
CA3DMS~\cite{ca3dms} &22.3&22.8&21.8 & &-&-&-\\
DiMP~\cite{dimp} &-&-&- & &54.3&70.3&73.1\\
ATOM~\cite{atom} &-&-&- & &50.5&59.8&68.8\\

    \bottomrule
    \end{tabular}
    }
  }
\label{tab:rgbd}
  \vspace{4mm}
\end{table}

\begin{table*}[t]
\caption{SOTA comparison on $\mathrm{DepthTrack}_{miss}$ \cite{depthtrack} for RGB-Depth, $\mathrm{LasHeR}_{miss}$\cite{li2021lasher} and $\mathrm{RGBT234}_{miss}$\cite{rgbt234} for RGB-Thermal, and $\mathrm{VisEvent}_{miss}$\cite{wang2023visevent} for RGB-Event.}
\label{tab-sota-miss}
\centering
\setlength{\tabcolsep}{3mm} 
\small 
\begin{tabular}{l|ccc|cc|cc|cc}
\toprule
\multirow{2}{*}{Method} 
& \multicolumn{3}{c|}{DepthTrack$_{miss}$~\cite{depthtrack}} 
& \multicolumn{2}{c|}{LasHeR$_{miss}$~\cite{li2021lasher}} 
& \multicolumn{2}{c|}{RGBT234$_{miss}$~\cite{rgbt234}} 
& \multicolumn{2}{c}{VisEvent$_{miss}$~\cite{wang2023visevent}} \\
\cline{2-4} \cline{5-6} \cline{7-8} \cline{9-10}
& F-score($\uparrow$) & Re($\uparrow$) & Pr($\uparrow$) 
& P($\uparrow$) & AUC($\uparrow$) 
& MPR($\uparrow$) & MSR($\uparrow$) 
& P($\uparrow$) & AUC($\uparrow$) \\
\midrule[0.5pt]
\textbf{FlexTrack} & \textbf{\textcolor{red}{57.8}} & \textbf{\textcolor{red}{56.1}} & \textbf{\textcolor{red}{59.6}} 
& \textbf{\textcolor{red}{65.1}} & \textbf{\textcolor{red}{52.3}} 
& \textbf{\textcolor{red}{84.1}} & \textbf{\textcolor{red}{62.6}} 
& \textbf{\textcolor{red}{72.8}} & \textbf{\textcolor{red}{55.0}} \\
\midrule[0.1pt]
STTrack \cite{sttrack} & 49.9 & 48.8 & 51.0 & 54.5 & 44.9 & 73.8 & 54.2 & 65.5 & 49.7 \\
SUTrack~\cite{chen2024sutrack} & 49.5 & 47.3 & 51.9 & 58.3 & 47.6 & 82.0 & 60.8 & 66.6 & 50.5 \\
SeqTrackv2~\cite{seqtrack} & 45.0 & 40.9 & 50.0 & 50.0 & 39.9 & 70.8 & 49.9 & 57.6 & 43.1 \\
SDSTrack~\cite{hou2024sdstrack} & 46.7 & 42.0 & 52.7 & 52.5 & 43.1 & 67.0 & 48.8 & 62.6 & 46.9 \\
IPT~\cite{ipt} & - & - & - & 61.7 & 46.4 & 82.0 & 59.4 & - & - \\
ViPT~\cite{vipt} & 44.4 & 40.5 & 46.6 & 40.1 & 34.0 & 52.4 & 39.4 & 57.2 & 43.2 \\
MCITrack~\cite{mcitrack} & 49.7 & 42.9 & 59.1 & 34.2 & 40.0 & 53.6 & 40.9 & 49.9 & 36.5 \\
\bottomrule
\end{tabular}
\end{table*}

\noindent \textbf{RGB-Event missing modality datasets:}
As illustrated in Tab.~\ref{tab-sota-miss}, our tracker achieves a P of 72.8\%, outperforming SUTrack~\cite{chen2024sutrack} by 6.2\%. Thanks to the video-level masking strategy, our model can capture more temporal information. This enables it to benefit from both the missing spatial information and the complete temporal information, even when the modalities are missing.

\vspace{1mm}
\noindent \textbf{RGB-Thermal missing modality datasets:}
To evaluate the scenarios involving missing modalities on the RGB-Thermal benchmark, we select two datasets, namely LasHeR~\cite{li2021lasher} and RGBT234~\cite{rgbt234}. As shown in Tab.~\ref{tab-sota-miss}, IPT~\cite{ipt} is a tracker specifically designed for the RGB-Thermal missing modality. It requires additional prompt parameters for tracking and needs to switch among three branches to handle different inference scenarios, including RGB-only, Thermal-only, and RGB-Thermal. In contrast, our tracker does not introduce new parameters for missing modality tracking. Instead, it only incorporates a novel video-level masking strategy. As a result, our tracker outperforms IPT~\cite{ipt} by 2.4\% on the LasHeR~\cite{li2021lasher} dataset and by 2.1\% on the RGBT234~\cite{rgbt234} dataset in terms of P.

\vspace{1mm}
\noindent \textbf{RGB-Depth missing modality datasets:}
We utilize DepthTrack~\cite{depthtrack} to evaluate the missing modalities in RGB-Depth tracking. The DepthTrack~\cite{depthtrack} dataset presents various challenges such as long-term tracking and similar object interference. In long-term videos, video information plays a crucial role in capturing the variations of the object to be tracked. However, once a modality is missing, some trackers fail to track the object due to the lack of long-term contextual information. However, with our masking strategy, our tracker can still benefit from the incomplete modalities in the past frames. Additionally, our tracker employs HMoE-Fuse to handle different cases of missing modalities, enhancing the tracker's flexibility in dealing with various missing modality scenarios. As shown in Tab.~\ref{tab-sota-miss}, our tracker significantly outperforms other trackers.

\subsection{Ablation Study}

In subsequent experiments, we conducted evaluations on LasHeR~\cite{li2021lasher} and VisEvent~\cite{wang2023visevent} datasets along with their missing-modality variants, using P as the primary metric to assess different methods.

\vspace{1mm}
\noindent \textbf{Video-level masking strategy:}
To enhance the robustness of our tracker, we propose a novel video-level masking strategy. To validate its effectiveness, we conducted experiments with and without the masking strategy, as shown in Table~\ref{tab:masking_variation}. For common multimodal datasets (e.g., LasHeR \cite{li2021lasher} and VisEvent  \cite{wang2023visevent}), the performance remains consistent when the masking strategy is removed. However, in scenarios with missing modalities (e.g., $
\mathrm{LasHeR}_{miss}$ and $\mathrm{VisEvent}_{miss}$), our masking strategy significantly improves performance compared to the baseline without masking. Specifically, we achieve a Precision (P) of 65.1 on $\mathrm{LasHeR}_{miss}$ and 72.8 on $\mathrm{VisEvent}_{miss}$, outperforming the no-masking variant by 3.8\% and 6.5\%, respectively.

Additionally, we compared our strategy with widely used methods, including random masking and VideoMAE \cite{tong2022videomae}. While these methods slightly outperform the no-masking baseline in missing modality scenarios, they lead to performance drops in complete datasets (e.g., LasHeR and VisEvent). For instance, random masking reduces the Precision on LasHeR from 77.3 to 77.0, and VideoMAE further degrades it to 74.3. 
This suggests that these methods disrupt the temporal continuity required for robust video tracking. In contrast, our video-level masking strategy preserves temporal context, ensuring stable performance across both complete and missing modality scenarios.

\begin{table}[t]
    \centering
    \caption{Performance comparison of different masking strategies.}
    \vspace{-3mm}
    \resizebox{\columnwidth}{!}{
    \begin{tabular}{c|cccc}
        \toprule
        & LasHeR & VisEvent & \text{LasHeR}$_{miss}$  &  \text{VisEvent}$_{miss}$  \\
        \midrule
        Ours & 77.3 & 81.4 & 65.1 & 72.8 \\
        \midrule
        w/o & 77.1 & 81.5 & 61.3 & 66.3 \\
        Random & 77.0 & 80.5 & 62.4 & 70.8 \\
        VideoMAE & 74.3 & 78.2 & 62.8 & 71.0 \\
        \bottomrule
    \end{tabular}
    }
    \vspace{2mm}
    \label{tab:masking_variation}
\end{table}

\vspace{1mm}
\noindent \textbf{HMoE-Fuse}
The HMoE-Fuse is a crucial component for the efficient fusion of diverse modalities. In Table~\ref{tab:HMoE-Fuse-component}, we present an ablation study examining the impact of various design choices on the performance of HMoE-Fuse. Specifically, "w/o gating" indicates the removal of the gating function, "w/o linear attn." refers to the absence of linear attention, and "w/o HMoE" denotes the removal of the HMoE module entirely. The experimental results demonstrate that eliminating these components leads to a significant decline in performance, highlighting their importance. To further validate the effectiveness of our HMoE, we replaced it with DeepSeekMoE\cite{dai2024deepseekmoe}. The results show that, for video-level tasks, HMoE-Fuse, which incorporates temporal-spatial information from video clips, outperforms DeepSeekMoE. This is attributed to the heterogeneous experts in HMoE-Fuse, which enable the tracker to have richer representations and better handle the complexities of video data. The superior performance of HMoE-Fuse underscores its advantage in leveraging diverse expert sizes to enhance tracking accuracy and robustness.

\begin{table}[t]
    \centering
        \caption{Ablation studies on HMoE-Fuse.}
                \vspace{-3mm}
    \resizebox{\columnwidth}{!}{
    \begin{tabular}{ c |c c c c}
        \toprule
       & LasHeR & VisEvent & \text{LasHeR}$_{miss}$  &  \text{VisEvent}$_{miss}$  \\
        \midrule
        Ours & 77.3 & 81.4  & 65.1 & 72.8  \\
                \midrule

        w/o gating  & 76.3 & 79.9  & 63.1  & 71.7  \\
        w/o linear attn.  & 75.4  & 78.2  & 61.8   & 69.9  \\
        w/o HMoE & 74.2  & 79.3  & 60.5   & 67.5  \\
        DeepseekMoE  & 75.0  & 81.2 & 62.1   & 68.7  \\

        \bottomrule
    \end{tabular}
    }
    \vspace{-2mm}
    \label{tab:HMoE-Fuse-component}
\end{table}

\begin{table}[t]
    \centering
        \caption{Ablation study on varying complexity.}
        \vspace{-3mm}
    \resizebox{\columnwidth}{!}{
    \begin{tabular}{ c |c c c c}
        \toprule
       & LasHeR & VisEvent & \text{LasHeR}$_{miss}$  &  \text{VisEvent}$_{miss}$  \\
        \midrule
        Ours & 77.3 & 81.4  & 65.1 & 72.8  \\
                \midrule
        512  & 76.1 & 80.9  & 62.7  & 71.7  \\
        64  & 76.4  & 80.1  & 63.8   & 70.6  \\
        4  & 77.0  & 80.8 & 64.1   & 71.4  \\
        \bottomrule
    \end{tabular}
    }
    \vspace{2mm}
    \label{tab:HMoE-Fuse-number}
\end{table}

\begin{figure*}[t]
\centering
\includegraphics[width=.95\linewidth,keepaspectratio]{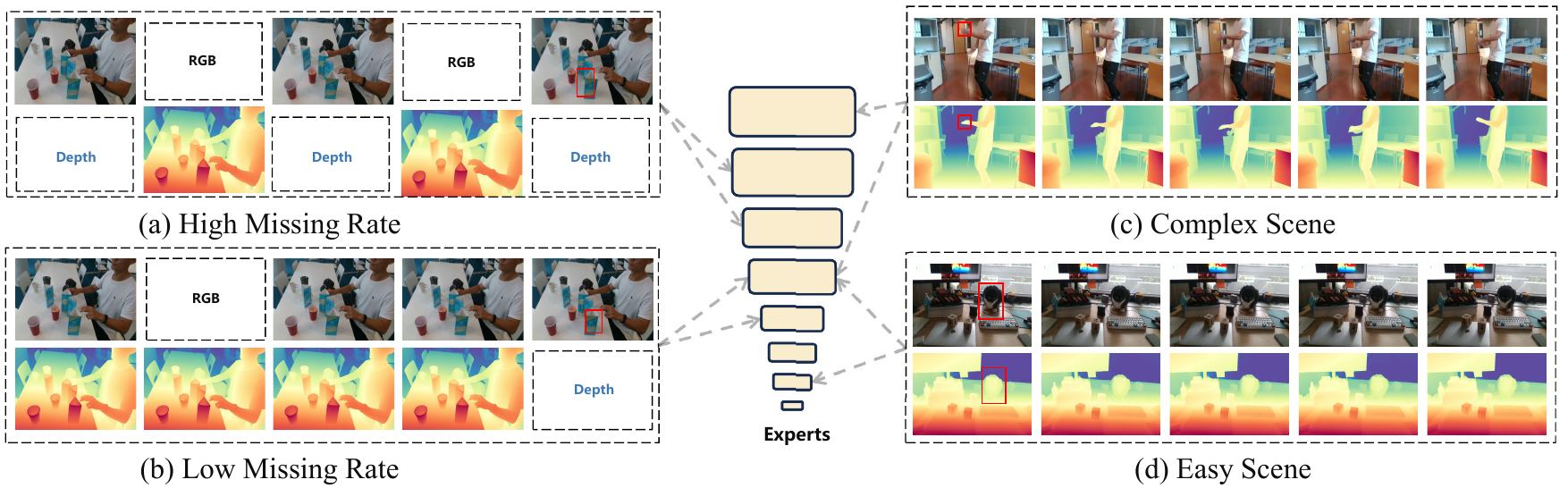}
\vspace{-3mm}
\caption{We are interested in the \textbf{following question}: \textit{With a higher missing rate, would a model favor a \textbf{lighter} expert or \textbf{a more complex} expert?} To investigate this, we analyze the impact of the missing modality rate on the same video clip, as shown in (a) and (b). When the missing rate increases (e.g., 50\% vs. 20\%), we observe that the model adaptively selects more complex experts with larger capacities to compensate for information scarcity through deeper reasoning. 
This insight leads us to a \textbf{second question}: \textit{Does this \textbf{behavior persist} in a full-modality setting?} To answer this, we conduct an analysis in (c) and (d), visualizing expert selection in a similar indoor scene under varying levels of difficulty—such as object size (e.g., a moving hand vs. a stationary cat). Our findings reveal that the model prioritizes high-capacity experts to enable fine-grained cross-modal fusion, effectively addressing the challenges posed by dynamic variations in the scene. The block size in the middle represents the hidden dimension of heterogeneous experts in the MoE architecture, where a larger dimension indicates higher complexity.}
\vspace{-2mm}
\label{fig:moe-selection}
\end{figure*}

In this paper, we employ the HMoE to achieve a more complex and nuanced representation. To validate the effectiveness of our proposed HMoE approach, we conducted experiments using both medium-sized and large-sized experts. As shown in Table~\ref{tab:HMoE-Fuse-number}, our HMoE consistently outperforms other configurations across various metrics. Compared to the homogeneous MoE models, the heterogeneous design of our MoE enables the model to achieve stronger expressive capabilities. This advantage is particularly evident under different missing rates and in challenging scenarios, where our model demonstrates superior performance.

\vspace{1mm}
\noindent \textbf{Visualization and Analysis:} To gain a deeper understanding of the behavior of our HMoE-Fuse module under varying degrees of missing modalities and full modalities, we conducted a visualization analysis of the selection  of experts with different missing rate and different full modalities scenes. The results are presented in Fig.~\ref{fig:moe-selection}.

From the visualizations, we observed that as the missing modality ratio increases, the model tends to select experts with larger hidden dimensions. This is primarily because, when a high proportion of modalities are missing, the model opts for larger experts to ensure precision through more complex reasoning. In contrast, with only a few modalities missing, the model exhibits higher confidence in its predictions and thus favors smaller experts for inference.

A similar pattern is evident in the full-modality scenario under complex conditions. For instance, in Fig.~\ref{fig:moe-selection} (c), the depth and feature differences between the two hands are minimal, making it challenging for the model to derive accurate tracking results. As a result, the model selects a larger expert to tackle the increased difficulty. Conversely, in Fig.~\ref{fig:moe-selection} (d), the distinct features of the ``cat" compared to its surroundings enable the model to easily pinpoint its location, allowing for the use of smaller experts during tracking.

\vspace{1mm}
\noindent \textbf{Takeaways:} Our HMoE framework offers a unique approach to interpretable multimodal fusion by explicitly linking the missing modality problem to complexity modeling — where \textit{more modalities reduce complexity}, and vice versa. We observe that, in simpler scenarios when more information is available, the model favors lighter experts, as the increased joint information entropy from the input enhances scene understanding. Our framework demonstrates that multimodality not only improves performance but also contributes to a more adaptive system. This insight is introduced and discussed for the first time in the context of multimodal video tracking, aligning with the fundamental principles of multimodal fusion.

\section{Conclusion and Future work}
In this work, we tackle the critical yet often understudied challenge of robust multimodal tracking under missing modality conditions. To address this, we propose a novel MoE design with heterogeneous experts, paired with a video-level mask training strategy. Additionally, our model can adaptively select experts of different scales during testing, depending on the extent of missing modalities. Experimental results show that our model excels across all leading benchmarks, with both complete and missing modalities.

While FlexTrack enhances multimodal tracking robustness, our future work will focus on integrating generative models for modality reconstruction and exploring lightweight deployment. Nonetheless, this work highlights the potential of heterogeneous expert systems in addressing the inherent unpredictability of real-world multimodal data.

\vspace{1mm}
\noindent \textbf{Acknowledgements.}
The authors sincerely thank the reviewers and all members of the program committee for their tremendous efforts and incisive feedback. 
This research was supported in part by the Alexander von Humboldt Foundation and in part by the Ministry of Education and Science of Bulgaria (support for INSAIT, under the Bulgarian National Roadmap for Research Infrastructure). C. Ma was supported in part by NSFC (62376156, 62322113).

{
    \small
    \bibliographystyle{ieeenat_fullname}
    \bibliography{main}
}

\end{document}